\documentclass[conference]{IEEEtran}
\IEEEoverridecommandlockouts

\usepackage{lscape}
\usepackage{array}
\usepackage{graphicx}
\usepackage{multicol}
\usepackage{multirow}
\usepackage{amssymb}
\usepackage{pifont}
\usepackage{tikz}
\usepackage{verbatim}
\usepackage{array}
\usepackage{longtable}
\usepackage{supertabular}
\usepackage{amsmath}
\usepackage{epstopdf}
\usepackage{verse}
\usepackage{cite}

\usepackage{color}
\usepackage{lettrine}
\usepackage{hyperref}
\usepackage{soul}
\usepackage{comment}
\usepackage{verbatim}

\usepackage{amsfonts}
\usepackage{flushend}
\def\BibTeX{{\rm B\kern-.05em{\sc i\kern-.025em b}\kern-.08em
		T\kern-.1667em\lower.7ex\hbox{E}\kern-.125emX}}
\usepackage{tikz}

\usepackage{amsmath}
\usepackage{pifont}

\DeclareMathOperator*{\argmaxA}{arg\,max} % thin space, limits underneath in displays

\usepackage{algorithm}
\usepackage{algorithmic}
\newcolumntype{a}{>{\columncolor{LightCyan}}c}

\makeatletter
\newcommand{\linebreakand}{%
\end{@IEEEauthorhalign}
\hfill\mbox{}\par
\mbox{}\hfill\begin{@IEEEauthorhalign}
}
\makeatother
\begin{document}
	
	\title{Compressive Features in Offline Reinforcement Learning for Recommender Systems\\
		
		\thanks{* Hung Nguyen and Minh Pham are co-first authors.}
	}

	\author{\IEEEauthorblockN{1\textsuperscript{st} Minh Pham \textsuperscript{*}}
		\IEEEauthorblockA{\textit{Computer Science and Engineering} \\
			\textit{University of South Florida}\\
			Tampa, USA \\
			minhpham@usf.edu}
		\and
		\IEEEauthorblockN{1\textsuperscript{st} Hung Nguyen \textsuperscript{*}}
		\IEEEauthorblockA{\textit{ Electrical Engineering} \\
			\textit{University of South Florida}\\
			Tampa, USA \\
			nsh@usf.edu}
		
		\linebreakand 
		\IEEEauthorblockN{2\textsuperscript{nd} Long Dang}
		\IEEEauthorblockA{\textit{Computer Science and Engineering} \\
			\textit{University of South Florida}\\
			Tampa, USA \\
			longdang@usf.edu}
		\and
		\IEEEauthorblockN{3\textsuperscript{rd} Jennifer Adorno Nieves}
		\IEEEauthorblockA{\textit{Computer Science and Engineering} \\
			\textit{University of South Florida}\\
			Tampa, USA \\
			jorgea1@usf.edu}
		
	}
	
	\IEEEpubid{\makebox[\columnwidth]{978-1-6654-3902-2/21\$31.00~\copyright2021 IEEE \hfill} \hspace{\columnsep}\makebox[\columnwidth]{ }}

	\maketitle	
	
	% ----------------------------------------------------------------------------

	\begin{abstract}
		
		In this paper, we develop a recommender system for a game that suggests potential items to players based on their interactive behaviors to maximize revenue for the game provider. Most of today's recommender systems in e-commerce and retail businesses are built based on supervised learning models and collaborative filtering, while our approach is built on a reinforcement-learning-based technique and is trained on an offline data set that is publicly available on an IEEE Big Data Cup challenge. The limitation of the offline data set and the curse of high dimensionality pose significant obstacles to solving this problem. Our proposed method focuses on improving the total rewards and performance by tackling these main difficulties. More specifically, we utilized sparse PCA to extract important features of user behaviors. Our Q-learning-based system is then trained from the processed offline data set. To exploit all possible information from the provided data set, we cluster user features to different groups and build an independent Q-table for each group. Furthermore, to tackle the challenge of unknown formula for evaluation metrics, we design a metric to self-evaluate our system's performance based on the potential value the game provider might achieve and a small collection of actual evaluation metrics that we obtain from the live scoring environment. Our experiments show that our proposed metric is consistent with the results published by the challenge organizers. We have implemented the proposed training pipeline, and the results show that our method outperforms current state-of-the-art methods in terms of both total rewards and training speed. By addressing the main challenges and leveraging the state-of-the-art techniques, we have achieved the best public leaderboard result in the challenge. Furthermore, our proposed method achieved an estimated score of approximately 20\% better and can be trained faster by 30 times than the best of the current state-of-the-art methods.  
		
	\end{abstract}
	\begin{IEEEkeywords}
		Recommender System, Reinforcement Learning, Big Data
	\end{IEEEkeywords}
	\IEEEpeerreviewmaketitle
	% ------------------------------------------------------------------------------------------

	\section{Introduction}
	\label{introduction}
	
	In the last decades, recommender systems have been deployed in many industrial applications in many different fields. They aim to suggest items that would maximize user satisfaction and the provider's revenue. They can be seen in various online services such as music and video services, gaming, online retail, restaurants, and online dating. They become more and more critical, especially when retail e-commerce is growing rapidly. In 2020, global retail e-commerce sales were 5.23 trillion USD, and they are expected to reach 6.54 trillion USD in 2022.  Such growth poses many challenges for recommender systems, such as the volume and velocity of data, item-wise interactions, sparse data, dynamic user preference, and scalability \cite{ricci2015recommender}. The majority of today's e-commerce and retail businesses build their suggesting systems by conducting supervised learning-based models and aim at optimizing clients' satisfaction in a greedy manner. Nevertheless, the item-by-item greedy recommendation approach is not well-suited to many real-world applications. In sequential recommendation situations, for example, traditional supervised learning methods frequently assume different stages in a session to be independent and therefore miss the oportunity to find the optimal strategy. Furthermore, the conversion rate of an item does not just depend on the item itself among the recommended items. For example, when similar but more expensive items surround an item, the likelihood of purchasing it increases, which is known as the decoy effect \cite{zhang2007agent}. However, the number of possible combinations of all items can be in the billions, which is an NP-hard problem that is understudied in traditional supervised learning \cite{zhu2014bundle}. Recent research has been adopting reinforcement learning to tackle this challenge in suggesting systems. As recent research displays, the recommendation can be described as a series of interactions between the user (environment) and the recommender system. Since reinforcement learning algorithms are naturally designed to maximize long-term rewards, explore the combinatorial space, and solve multi-step decision-making challenges, utilizing reinforcement learning to design recommender systems is an interesting contribution.
	
	In this paper, we tackle the challenge of designing a recommender system for the trading system in an online game. The recommender system would suggest items to players that are suitable for their playstyle and stage in the game and maximize the revenues for the game provider. This problem was proposed by the FUXI AI Lab, Netease, and was organized in the frame of the IEEE Big Data 2021 Cup. In this challenge, participants are given an offline data set with more than 250,000 playing sessions, 381 items, and approximately 40,000 users \cite{2021RL4RS}. There are three stages in each episode of the game, and in each stage, a user is recommended three items. The goal of the recommender system in this game is to maximize revenues earned from the nine recommended items. The features in the data set include user portraits, clicking history, their stages in the game, item features, and which items were purchased by each user. To tackle this problem, we proposed a design of a recommender system where roughly 400 features are examined and extracted by an unsupervised algorithm. This information is then used to cluster the state of the environment, and parallel groups of reinforcement models are trained on the clustered information. By implementing this strategy, we helped the reinforcement learning models learn the most important features from the environment, thus providing better-suited recommendations based on user portraits and their stages in the game. After that, by utilizing the Bellman equation as a simple value iteration update \cite{richard1954theory}, our parallel reinforcement learning models can find sets of items to generate maximum revenues.
	
	Our contributions in this work can be described as follows.
	\begin{itemize}
		\item We designed a flexible and high-performance reinforcement learning system based on feature extraction, clustering, and parallel training of Q-learning models
		\item We implemented the proposed system with an optimized performance by using parallelism on CPU and GPU.
		\item We compared the results of our system with other state-of-the-art Reinforcement Learning approaches and showed that our method could generate higher revenues than other methods. Furthermore, our system can be trained faster than the other methods by up to 24 times.
	\end{itemize}
	
	The remainder of this  paper is organized as follows. In Section \ref{sec:background}, we describe the problem in detail. Section \ref{sec:related_work} briefly summarizes existing state-of-the-art solutions. Our solution is introduced in Section\ref{sec:method}. Section \ref{sec:Implementation} describes the implementation details of our proposed solution. Section \ref{sec:Results} presents our results and the comparisons to other techniques.
	%%%%%%%%%%%%%%%%%%%%%%%%%%%%%%%%
	
	\section{Problem Description}
	\label{sec:background}
	
\begin{figure}[h!]
		%[trim=left bottom right top, clip]
		\includegraphics[width=\linewidth, trim=280 150 300 140,clip]{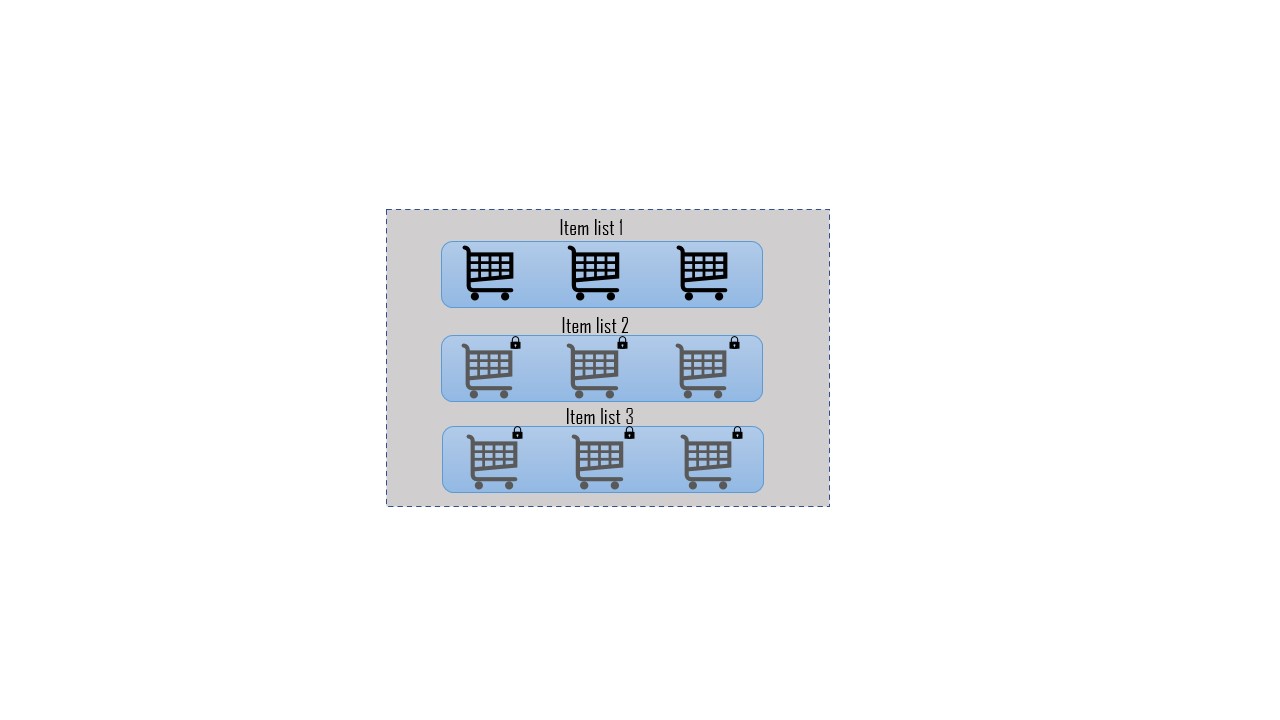}
		\caption{Item recommendation system.}
		\label{fig:9items}
	\end{figure}

	In each user session, the goal is to recommend nine items such that the total revenue from purchased items is maximized. As shown in Figure \ref{fig:9items}, the recommender system needs to respond to each client request with three-item lists (3 items per list), and the next item lists cannot be purchased until all three items on the current list are purchased. The users' purchasing behavior is influenced by the items on the next lists as well as the current item list. This challenge can be framed as a multi-stage decision-making problem in two ways. In the first way, there are 9 stages with one item recommended in each stage. In the second way, there are 3 stages with three items recommended in each stage.
	
	A live interactive training environment was not provided to the participants. Instead, an offline training data set was provided. The training data set is gathered over a three-month period and includes a few sales initiatives. Essentially, the environment is a test system on clients' purchase data and is built using data 10 times the size of the data being generated. The test environment is the same as it was when the recommendation was evaluated, and it is not available to participants of the competition. The training data set includes information about user sessions and items. In each user session, the provided information includes a timestamp, the user's click history on all items, ten user portraits features, the nine items recommended by a recommender system, and a list of nine binary labels indicating whether the items were purchased. For each item, information about the item's content features, price, and location are provided. An item can only be recommended on a list whose location matches the item's location. For example, an item with location 1 can only be recommended on the first item list, i.e., only position 1, 2, or 3 of the nine recommended items. In other words, an item with location 1 can only be placed on the first row of Figure \ref{fig:9items}.
	
	Lacking a live environment is a common situation in real-life applications of recommender systems due to various reasons such as security, user privacy, high cost for maintaining a live environment \cite{levine2020offline}. However, the lack of a live environment for training and testing a recommender system creates many difficult challenges for designing a reinforcement learning-based approach, such as issues with bootstrapping from out-of-distribution actions and overfitting. These problems are caused by erroneously optimistic value estimates for sample states that fall outside of the scope of the offline training data set \cite{ma2021conservative}. In Section \ref{sec:method}, we will describe our approach for overcoming the difficulties of training from an offline data set.
	
	%%%%%%%%%%%%%%%%%%%%%%%%%%%%%%%%%%
	
	\section{Related Work}
	\label{sec:related_work}
	
	In this section, we describe the current state-of-the-art development in the literature. We use the notations from Hasselt et al. \cite{van2016deep} and Fujimoto et al.\cite{fujimoto2019off}.In reinforcement learning, an agent interacts with its environment, typically assumed to be a Markov decision process (MDP). 
	
	In \cite{fujimoto2019off}, the MDP process is formally defined by a tuple ($S, A, p_M, r, \gamma$), where each component of the tuple is listed below:
	
	\begin{itemize}
		\item $S$ are state spaces,
		\item $A$ denotes action spaces,
		\item $p_M$($s^\prime|s, a$) is a transition dynamic,
		\item $r(s, a, s^\prime) \in R$ represent a reward which an agent receives when performing action $a$ in state $s$ and ending at the state $s^\prime$,
		\item $\gamma \in (0, 1)$ is a discount factor.
	\end{itemize}
	
	%%%%%%%%%%%%%%%%%%%%%%%%%%%%%%%%%%
	At each discrete time step, the objective of the agent is to maximize the expected sum of discounted rewards when taking action $a_i$ in state $s_i$ which is defined as follows.
	
	\begin{equation}\label{ep:defR_t}
	R_t = \sum_{i=t+1}^{\infty} \gamma^{i}r(s_i, a_i, s_{i+1})
	\end{equation}
	
	%%%%%%%%%%%%%%%%%%%%%%%%%%%%%%%%%%
	
	The agent selects actions with respect to a policy $\pi:S \rightarrow A$ which exhibits a distribution $\mu^{\pi}(s)$ over the states $s \in S$ visited by the policy. Under the given policy $\pi$, the true value of an action $a$ in a state $s$ is $Q^{\pi}(s, a) = \mathbb{E}_{\pi}[R_t|s ,a]$. The corresponding action value can be computed through the Bellman operator $T^{\pi}$: $T^{\pi}Q(s, a) = \mathbb{E}_{S_{\prime}}[r + \gamma Q^{*}(s^{\prime}, \pi(s^{\prime})]$ where $Q^{*}(s^{\prime}, a^{\prime}) = \max_{\pi} Q^{\pi}(s^{\prime}, a^{\prime}) $. The Bellman operator states that the estimated long-term reward for a given action can be determined by the immediate reward from the given action plus the expected reward from the best future action taken at the next state. By picking the highest valued action in each state, an optimal policy is easily found from the optimal values.
	
	%%%%%%%%%%%%%%%%%%%%%%%%%%%%%%%%%%%%%%%
	\subsection{Q-learning algorithm}
	Since most interesting problems such as playing modern video games have nearly infinitely large number of possible states, it is extremly difficult to compute all action values in all states separately using the Bellman operator. Q-Learning proposed by Watkins \cite{watkins1989learning} attempts to learn estimates of the optimal action (reward) values called Q-values when taking an action $a$ in a particular state $s$ via a parameterized value function. Watkins et al. \cite{watkins1989learning} proposed that we can learn the  Q-values of all actions in any number of possible states using the parameterized value function $Q(s, a;\theta_t)$, in which we would like to find the weights $\theta$ that updates the value of the function $Q(s, a; \theta_t)$ towards a target value $Y^{Q}_{t}$. The weights can be found using stochastic gradient descent algorithm. Therefore, the standard Q-learning update for the parameters after taking action $A_t$ in state $S_t$ and observing the immediate reward $R_{t+1}$ and resulting state $S_{t+1}$ is then: 
	\begin{align}\label{ep:defQ}
	\theta_{t+1} = \theta_{t} + \alpha (Y^{Q}_{t} - Q(S_t, A_t; \theta)) \triangledown_{\theta_t}Q(S_t, A_t; \theta)
	\end{align} where $\alpha$ is a scalar step size and the target $Y^{Q}_{t}$ is determined as follows.
	
	\begin{equation}\label{ep:defTQ}
	Y^{Q}_{t} \approx R_{t+1} + \gamma \max_{a} Q(S_{t+1}, a; \theta_{t})
	\end{equation}
	
	%%%%%%%%%%%%%%%%%%%%%%%%%%%%%%%%%%%%%%%
	\subsection{Deep Q-Network (DQN)}
	To make Reinforcement Learning more applicable to  real-world problems, Mnih et al. \cite{mnih2015human} proposed using a deep convolutional neural network (CNN) in combination with Q-learning.
	
	%%%%%%%%%%%%%%%%%%%%%%%%%%%%%%%%%%%%%%%%%%%%%%%%
	A DQN is a multi-layered convolutional neural network that maps a given state $s$ to a vector of action values $Q(s, ., \theta$), where $\theta$ are the parameters of the network. The Q-network can be written as a function $f: \{0, 1\}^n \rightarrow R^m$ that maps an input state $s \in \{0, 1\}^n$ to an output $y \in R^m$ where $n$ is the number of states, $m$ is number of actions, and $R$ is a set of real numbers.
	%%%%%%%%%%%%%%%%%%%%%%%%%%%%%%%%%%%%%$$$$$$$$$$
	Two important ingredients of the DQN algorithm are the use of a second target network and the use of experience replay \cite{van2016deep}. 
	%%%%%%%%%%%%%%%%%%%%%%%%%%%%%%%%%%%%%%%%%%%%%%%%
	We utilize the target network's parameter $\theta ^{-}_{t}$ to calculate the target action values $Y^{DQN}_{t}$. The target network with parameters $\theta ^{-}_{t}$ has the same architecture as the Q-network but with parameters being copied every $\tau$ steps from the Q-network, so that $\theta ^{-}_{t}$ = $\theta_t$, and kept fixed on all other steps. The target used by DQN is defined as follows.
	
	\begin{equation}\label{ep:defDQN}
	Y^{DQN}_{t} \approx R_{t+1} + \gamma \max_{a} Q(S_{t+1}, a; \theta ^{-}_{t})
	\end{equation}
	
	%%%%%%%%%%%%%%%%%%%%%%%%%%%%%%%
	The second addition to considerably increase the DQN algorithm's performance is experience replay \cite{lin1992self}. The main idea of the experience replay is that we can store agents' observed transitions and uniformly sample batches of them to train the CNN.
	
	%%%%%%%%%%%%%%%%%%%%%%%%%%%%%%%%%%%%%%%
	\subsection{Double Deep Q-Network (Double DQN)}
	The main motivation behind Double DQN proposed by Hasselt et al. \cite{van2016deep} is that the Q-network often overestimates the action values. In order to reduce overestimations, instead of including a maximization step when computing the target Q-values in the target network, the authors propose using the Q-network in the DQN architecture to choose an action and utilize the target network to generate the target Q-values for that action. By decomposing the max operation in the target into the action selection and action evaluation, the overestimation is substantially reduced. Below is the new Double DQN equation for updating the target value:
	
	\begin{small}
		\begin{align}
		\label{ep:defDoubleDQN}
		Y^{DoubleDQN}_{t} \approx R_{t+1} + \gamma Q(S_{t+1}, \argmaxA_a Q(S_{t+1}, a; \theta _{t}), \theta ^{-}_{t})
		\end{align}
	\end{small}
	%%%%%%%%%%%%%%%%%%%%%%%%%%%%%%%%%%%%%%%
	\subsection{Batch Constrained deep Q-learning (BCQ)}
	In 2019, Fujimoto et al. \cite{fujimoto2019off} proposed an offline (or "batch") RL method which aims to train an agent to learn from large offline data sets, but without any interaction with the environment. In real-world problems, an environment interaction may be expensive, unsafe and time consuming \cite{wei2021boosting}. In this work, the main idea is to run normal Q-learning. Instead of evaluating the max action value over all potential actions shown in Equation (\ref{ep:defTQ}), we only want to examine actions $a^\prime$ that $(s^\prime,a^\prime)$ really existed in the batch of data by removing actions that are unlikely to be chosen by the behavior policy in each batch $\pi_b$ \cite{fujimoto2019benchmarking}. 
	%%%%%%%%%%%%%%%%%%%
	Firstly, BCQ utilizes a state-conditioned generative model $G_{\omega}(s)$ which given the state as input, produces actions that are likely to be selected from the batch.
	%%%%%%%%%%%%%%%%%%%%%%%%%%%
	Secondly, BCQ contains a perturbation model $\xi_{\phi}(s, a)$, which further modifies the actions within a set range $[-\Phi, \Phi]$ and trains the perturbation model using the deterministic policy gradient \cite{silver2014deterministic}.
	%%%%%%%%%%%%%%%%%%%%
	Finally, the authors use a weighted version of Clipped Double Q-learning
	%%%%%%%%%%%%%%%%%%%
	At test time, the authors sample $N$ actions via the generator, perturb each, and pick the action with the highest estimated Q-value. Therefore, the policy is defined as below: 
	\begin{equation}\label{ep:defBCQ}
	\pi(s) = \max_{a_i + \xi_{\phi}(s, a_i)} Q(s, a_i + \xi_{\phi}(s, a_i) ; \theta)   
	\end{equation} 
	where ${[a_i \sim G_{\omega}(s)]}_{i=1}^{N}$.
	%%%%%%%%%%%%%%%%%%%%%%%%%%%%%%%%%%%%%%%%%%%%
	\subsection{Conservative Q-learning (CQL)}
	Existing offline RL methods suffer from a major challenge related to the distribution shift between the behaviour policy dataset and the learned policy \cite{kumar2020conservative}. This limitation results in an overestimating Q-values for out-of-distribution states and actions. Therefore, poor actions are selected \cite{daylearning}. 
	%%%%%%%%%%%%%%%%%%%%%%%%%%%%%%%%%%%%%%%%%%%%%%%%%%%%%%%%%%%%%%%%%%%
	The CQL \cite{kumar2020conservative} algorithm addresses this overestimation by ensuring conservative approximation of the Q-values via penalizing Q-values for out-of-distribution states and actions \cite{daylearning}. As a result, the learned policy will be trained to keep closer to the known behaviors in the behaviour data rather than incorrectly favoring inflated values. 
	%%%%%%%%%%%%%%%%%%%%%%%%%%%%%%%%%%%%%%%%%%%%%%%%%%%%%%%%%%%%%%%%%
	To achieve such a lower-bound on the true Q-values, CQL uses an objective function $J_{CQL}(\theta)$ which depends on the choice of action distribution and is defined as follows.
	
	\begin{small}
		\begin{align}\label{ep:defCQL}
		J_{CQL}(\theta)& = \min_{Q} \mathbb{E}_{s \sim D} \left[\log \sum_{a} \exp{Q(s,a)} - \mathbb{E}_{a \sim \hat{\pi}_{\beta}(a|s) }{{Q(s,a)}} \right] \nonumber \\
		&+ \frac{1}{2}\mathbb{E}_{s, a, s^{\prime} \sim D} \left[r(s,a) + \gamma \mathbb{E}_{\pi}[\hat{Q}(s^{\prime}, a^{\prime})] - Q(s, a)^{2} \right]
		\end{align}
	\end{small}
	where $(s, a, s^\prime)$ represents state, action, next state tuples; $r(s, a)$ is the reward; $\hat{\pi}_{\beta}(a|s)$ defines the behaviour policy which generates the dataset $D$, and $\hat{Q}(s^{\prime}, a^{\prime})$ is the Q-function approximator \cite{daylearning}. 
	%%%%%%%%%%%%%%%%%%%%%%%%%%%%%%%%%%%%%%%%%%%%%%
	\section{Methodology}
	\label{sec:method}
	
\begin{figure*}[h!]
		%[trim=left bottom right top, clip]
		\centering
		\includegraphics[width=0.75\linewidth, trim=280 350 150 140,clip]{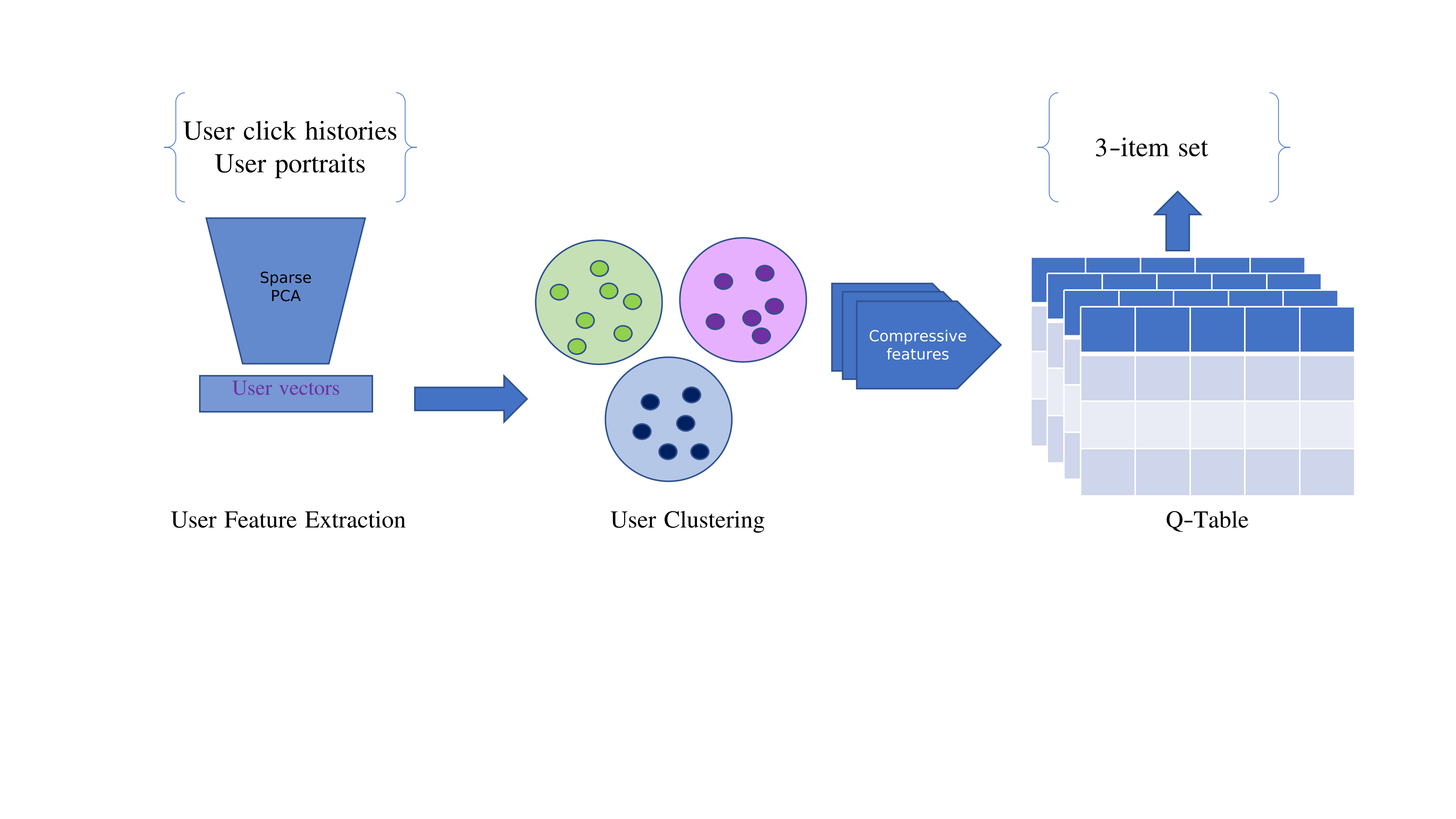}
		\caption{ComFOR's learning framework inlcuding preprocessing data, clustering and Q-learning. }
		\label{fig:method}
	\end{figure*}

	We propose a training pipeline as presented in Figure \ref{fig:method}. In this pipeline, we first perform feature extraction on the high-dimensional state space to reduce the problem of high dimensionality and retain features that are important to our reinforcement learning models. Next, we use the extracted features to cluster users. Finally, samples from grouped users are given to the parallel Q-Learning models according to their group membership. The Q-Learning models are updated by following the Bellman equation \cite{sutton2018reinforcement}.
	
	%%%%%%%%%%%%%%%%%%% Jing's method use additional data for training
	\subsection{Data Preprocessing}
	\label{sec:preprocessing}
	Preprocessing data involved parsing two text files: one that contains the user sessions and one that contains item information, as subsequently described. In the text file that contains the user sessions, each line contains a user ID, the user's click history (i.e., the items which have been previously clicked by the user), ten user portrait features, the nine items proposed to the user by a recommender system, nine binary labels indicating whether the user purchased each item, and a timestamp for the session. In the text file that contains item information, each line contains item ID (which can be matched to the first file), five-item content features, the item's price, and the item's location. An item can only be recommended on a list whose location matches the item's location. For example, an item with location 1 can only be recommended on the first item list, i.e., only position 1, 2, or 3 of the nine recommended items. In other words, an item with location 1 can only be placed on the first row of Figure \ref{fig:9items}.
	
	As mentioned in Section \ref{sec:related_work}, each training record for all reinforcement learning methods is a tuple of ($S, A, p_M, r, \gamma$), where $S$ defines state spaces, $A$ denotes action spaces, $p_M$($s^\prime|s, a$) denotes the next state given the current state and the taken action, $r$($s, a$) represents the reward, and $\gamma \in (0, 1)$ represents the discount factor. In this problem, we use a combination of user portrait features and user click history to represent states. There are 10 user portrait features and 381 one-hot-encoded user click history features, but we use a feature extraction method to reduce the number of features representing states (details in sub-section \ref{sec:PCA}). There are three ways to construct the action space: a nine-step decision-making problem with each action being an item ID, a three-step decision-making problem with each action being a set of three-item IDs, and a one-step decision-making problem with each action being a set of nine-item IDs. The nine-step setup has a significant disadvantage of not being able to hold multi-item interaction information. The one-step set up has a huge action space with ${381 \choose 9} = 4.24*10^{17}$ possible actions. Therefore, we chose the three-item setup because it allows us to maintain multi-item interaction while having a reasonable action space with ${381 \choose 9} = 4.24*10^{17} \approx$ 9 million possible actions. The next stage $p_M$($s^\prime|s, a$) can be determined as being terminated if the user does not purchase all three recommended items or being the next incremented step with the same user features. Reward $r$ is defined as the total price of the purchased items among the three recommended items.
	
	\subsection{Feature Extraction}
	\label{sec:PCA}
	
	As described in the previous section, the state space contains 10 user portrait features and 381 one-hot-encoded user click history features. Such a sparse state space may pose significant problems for deep learning algorithms \cite{krishnan2018challenges}. For example, our training data set may not have enough coverage on the entire feature space and is, therefore, more likely to have missed many maxima and minima of the loss function. Therefore, it is ideal to extract only the important features to reduce the dimensionality of the state space. Principal Component Analysis (PCA) \cite{wold1987principal} and AutoEncoder (AE) \cite{kramer1991nonlinear} are two common and standard methods for feature extraction and didimensionsssreduction. However, one drawback of these two techniques in dealing with sparse data is that they use all of the original features to compute the output features, which may introduce unnecessary noise from unimportant and rare features \cite{johnstone2009consistency}. To tackle this problem, Sparse PCA \cite{zou2006sparse} and Sparse AutoEncoder \cite{makhzani2013k} were introduced. It has been shown that the classical PCA can retain consistency when the number of features is much larger than the number of data points, a situation where the classical PCA generates large variance \cite{johnstone2009consistency}. We utilize these two methodologies to generate a state space with smaller dimensions from the training data set. The number of dimensions is selected through a cross-validation process by using our proposed scoring system as described in Section \ref{sec:metric}.
	
	\subsection{Clustering}
	\label{sec:clustering}
	One of the major drawbacks of the classical Q-Learning approach is that its memory requirement grows linearly with a product of the size of the state space and the size of the action space. Therefore, when the state space or the action space gets large, it is preferable to use Neural-Network-based approaches to approximate the quality value given a tuple of (state, action). In this problem, we have a large state space with high dimensionality. Even after dimension reduction, the state space is still infinitely large. To tackle this problem, we use clustering techniques to divide the state space into regions. Within each region of the state space, we construct sub-state space that only depends on the step of the item list. This means that all users who fall into a region of the state space would share the same Q-table and would be given the same action policy. The intuition behind this design is that we can divide users into groups based on their user portraits and click history and that users within a group are likely to react similarly to the same action policy. This kind of approach has been popular in human resources management and investment \cite{bauckhage2014clustering,hui2015spatial}.
	
	We utilized K-Means clustering and DBSCAN, the two standard approaches that have decent performance. The parameters on each approach are tuned, and the best approach is chosen based on our proposed scoring system as described in Section \ref{sec:metric}.
	
	\subsection{Q-Learning}
	\label{sec:QLearning}
	Q-learning is a model-free reinforcement learning algorithm that may be used to memorize the value of an action in a certain state \cite{sutton2018reinforcement}. It does not require a model of the environment and can handle stochastic interactions and rewards without modifications. Starting from any state, Q-Learning discovers an optimal solution for any finite Markov Decision Process in the sense of maximizing the expected total reward throughout any and all progressive stages. Given an infinite training time and a partly-random policy, Q-Learning may identify an optimal action-selection policy for every given Markov Decision Process. Q-Learning relies on storing in memory a Q-Table with size (Number of states) $\times$ (Number of actions). At initialization, all values in the Q-Table are initialized to 0. For each tuple of ($S, A, p_M, r, \gamma$) in the training data set, the Q-Table is updated according to the Bellman equation \cite{sutton2018reinforcement}: 
	
	\begin{small}
		\begin{align}
		\label{ep:QTable}
		Q^{new}(S,A) =& Q(S,A) +  \alpha (r + \gamma max(Q(p_M, a) - Q(S,A))
		\end{align}
	\end{small}
	
	where $\alpha$ is the learning rate. For all final states $S_f$, $Q(S_f,a)$ is never updated but is set to the reward value $r$ observed for the final state $S_f$.
	
	\subsection{Evaluation metric}
	\label{sec:metric}
	One key challenge in training offline reinforcement learning models is that the models often overestimate the reward values when calculating samples that are under-represented in the training data set \cite{levine2020offline}. There are two approaches to tackle this problem. The first approach is to restrict the recommended actions to the ones that are close to the training data set, such as the BCQ model proposed by Fujimoto et al. \cite{fujimoto2019off}. The second approach is to collect the training data set by running an online reinforcement learning model that has good coverage for all scenarios \cite{jin2021pessimism}. Our proposed method inherently follows the first approach. By ensuring that each user cluster has a sufficient amount of samples, we can guarantee to restrict the recommended actions in each cluster to the ones that are well represented because they are the ones with the highest Q values in the Q-Tables. Since the scoring metric is not disclosed to participants in this competition, we propose the following evaluation metric to evaluate the quality of a model quickly. By splitting the training data set into a train set and a validation set, we can train any model on the train set and use the model to make recommendations on the validation set. To evaluate our method performance, we design a metric that reflects potential rewards based on the price of recommended items and the user's actual purchased items. Our method is then evaluated on the testing set, which  consists of  20\% of the entire data set. The output of our metric is a weighted score that evaluates the total expense of a user for buying items in the game at different steps. Each 3-items list is a step, and the score in the higher steps should be put higher weights. This is an appropriate strategy since the game provider will benefit the most if the player survives through the end of the game. As the game rules restrict certain items on each step, if the items are not allowed to suggest on a particular step, their values will be zero. All recommended items in each step that are eventually purchased by the players will be counted in our final score. Our metric can be described in the algorithm as follows.
	
	\begin{algorithm}[ht]
		\caption{Algorithm to Calculate metric}
		\begin{flushleft}
			\textbf{Input}: Recommended Items {R}, Purchased Items {P}, Items' Prices , Restricted Items.\\
			\textbf{Parameter}: 
			$W_{st}$  is step weights\\
			N : number of testing samples\\
			\textbf{Output}: Score\\
			\begin{algorithmic}
				\FOR {st in steps }
				\STATE Items $\leftarrow$ i in ${R} \cup{P} $\\
				\FOR  {it in purchased Items and item in restricted items}
				\STATE $Value_{st} \; += it * Price_{it}$
				\ENDFOR
				\ENDFOR
				\STATE $Score = \frac{1}{N} \sum_{st} w_{st}*value_{st}$
				\RETURN 
			\end{algorithmic}
		\end{flushleft}
		\label{alg:COMFOR}
	\end{algorithm}

	The proposed evaluation metric can be used to compare the models and to tune parameters. As shown in Section \ref{sec:Results}, our evaluation metric are a fair representation of the real metric used by the competition organizer.
	
	%%%%----

	%-------------------------------------------------------------------------------------------
	
	% Factors are set before the attack begins, single machine with GPU,
	% DNN vs CNN
	% Complexity at the layer level
	% Computing – distributed
	% Gradient based attacks
	% Focus on data parallelism – centralized with a parameter server, synchronous ( best performance but slowliest, 
	
	\section{Implementation Details} 
	\label{sec:Implementation}
	We have implemented our proposed method in Python. For the feature extraction part, we utilized the library Scikit-Learn's SparsePCA method and the Sparse AutoEncoder implementation by Makhzani and Frey \cite{makhzani2013k}. For the clustering part, we utilized Scikit-Learn's KMeans and DBSCAN methods. Parameters for these methods are selected through cross-validation by using the proposed metric in Section \ref{sec:metric}. We have implemented the Q-Learning models for high-performance scaling on GPU. First, we initialized all the Q-Tables on the GPU's global memory. For each cell on the tables, we created an exclusive lock to prevent multiple threads from updating a cell at the same time. Since all tuples ($S, A, p_M, r, \gamma$) from the train set can be processed and used to update the Q-Tables independently, processing an entire batch is an embarrassingly parallel problem where we let one thread process the computation for a tuple and then atomically update the cells with the use of the exclusive locks. A parallel implementation on CPUs was also programmed in a similar manner.
	
	\section{Results} 
	\label{sec:Results}
\begin{figure}[h!]
		%[trim=left bottom right top, clip]
		\includegraphics[width=\linewidth, trim=0 0 0 0,clip]{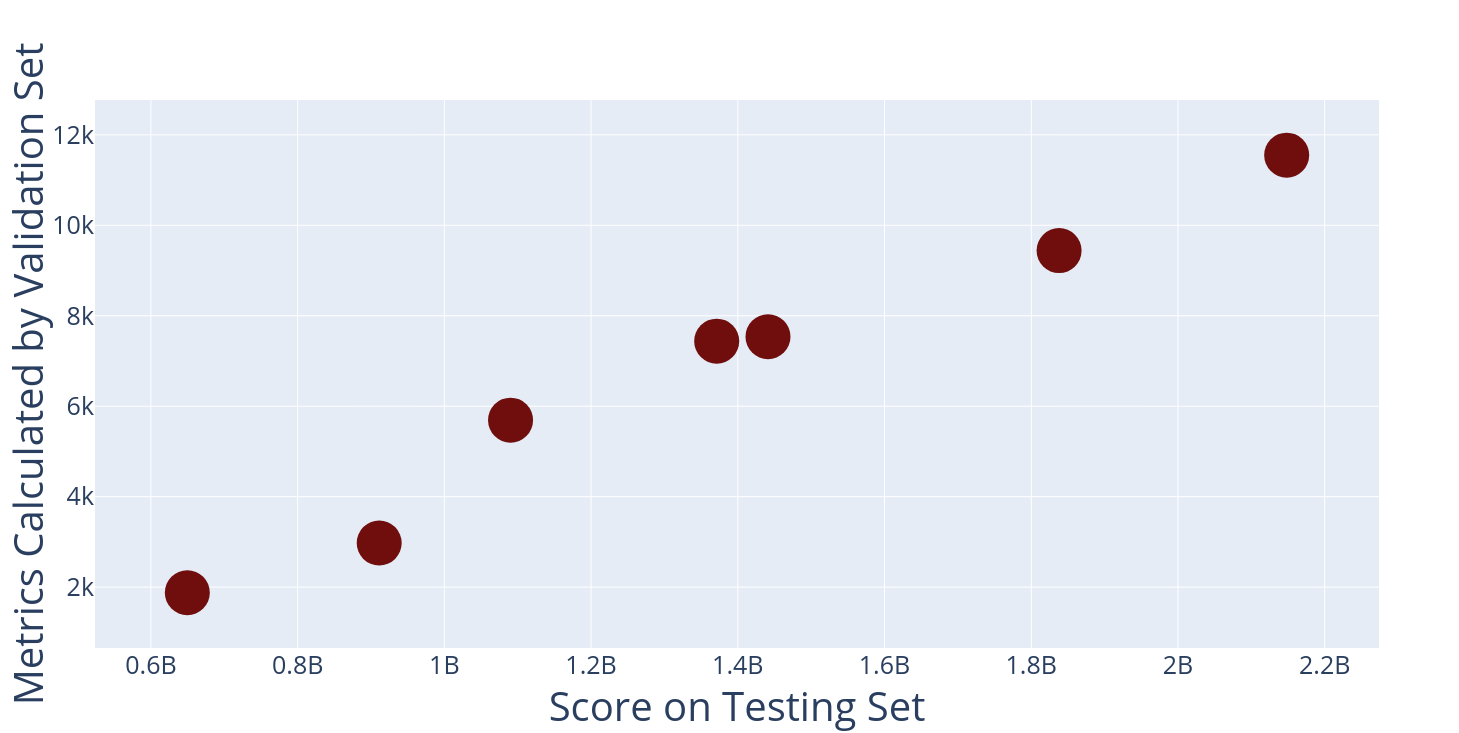}
		\caption{Our proposed metric versus leaderboard results.}
		\label{fig:metricResults}
	\end{figure}

	First, we show that our proposed metric is a good representation of the private metric used by the competition organizers. First, we split the training data set into an 80\% set for training and a 20\% set for validation. Then, the metric proposed in Section \ref{sec:metric} is calculated on the validation set. Finally, we use the trained model to make recommendations on the testing data set provided by the organizers and obtain the score from the organizers' scoring system. We repeat this process with the four deep reinforcement learning described in Section \ref{sec:related_work} and our proposed method with different sets of parameters (number of extracted features, clustering parameters, and learning rate for Q-learning). The resulted scores on the testing set and our calculated metric are plotted in Figure \ref{fig:metricResults}. Since the organizers allow only one query per day to the scoring system for calculating scores on the testing set, we could only obtain a few data points.
	
	Figure \ref{fig:metricResults} shows that our proposed metric is consistent in representing the organizer's calculated metric on the testing set.
	
\begin{figure}[h!]
		%[trim=left bottom right top, clipS
		\includegraphics[width=\linewidth, trim=0 0 0 0,clip]{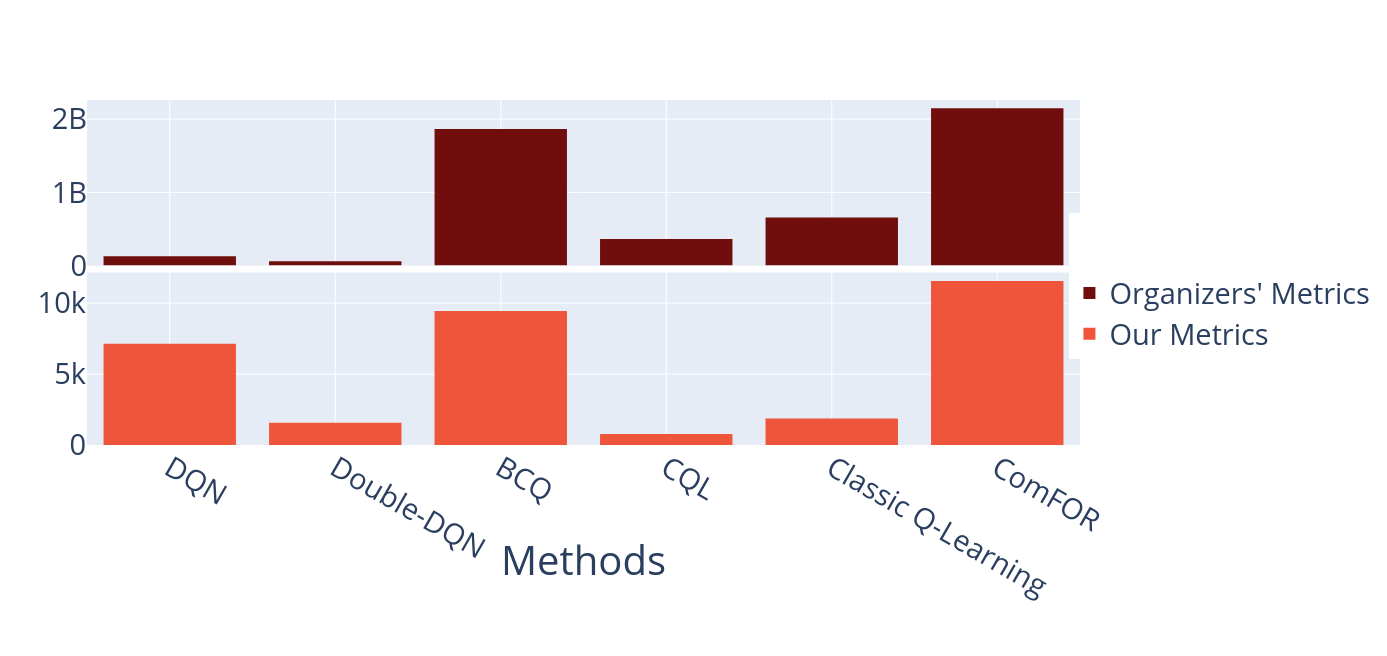}
		\caption{Results from different methods evaluated by our metric and the organizers' metric.}
		\label{fig:Results_Methods}
	\end{figure}

	Next, we show that our proposed method outperforms the four deep reinforcement learning methods described in Section \ref{sec:related_work} by using both our proposed metric and the organizers' scoring system. The four deep reinforcement learning methods are Deep Q-Network (DQN), Double Deep Q-Network (Double DQN), Batch Constrainted Deep Q-Learning (BCQ), and Conservative Q-Learning (CQL). We used the Python package d3rlpy for the implementation of these methods. In addition, we reproduced the classical Q-Learning method and applied it to this problem. Parameters are tuned by using cross-validation with our proposed metric, and only the best parameters for each method are reported. The results of DQN, Double DQN, BCQ, CQL, Classical Q-Learning, and ComFOR are presented in Figure \ref{fig:Results_Methods}. 
	
	In addition, the organizers released the official baseline scores as follows. By using logged offline actions, the optimal score achievable is 770,378,225. By using a Long-Short-Term-Memory environment simulator in combination with Deep Deterministic Policy Gradient, the optimal score achievable is 1,033,481,948.
	
\begin{figure}[h!]
		%[trim=left bottom right top, clip]
		\includegraphics[width=\linewidth, trim=0 0 0 0,clip]{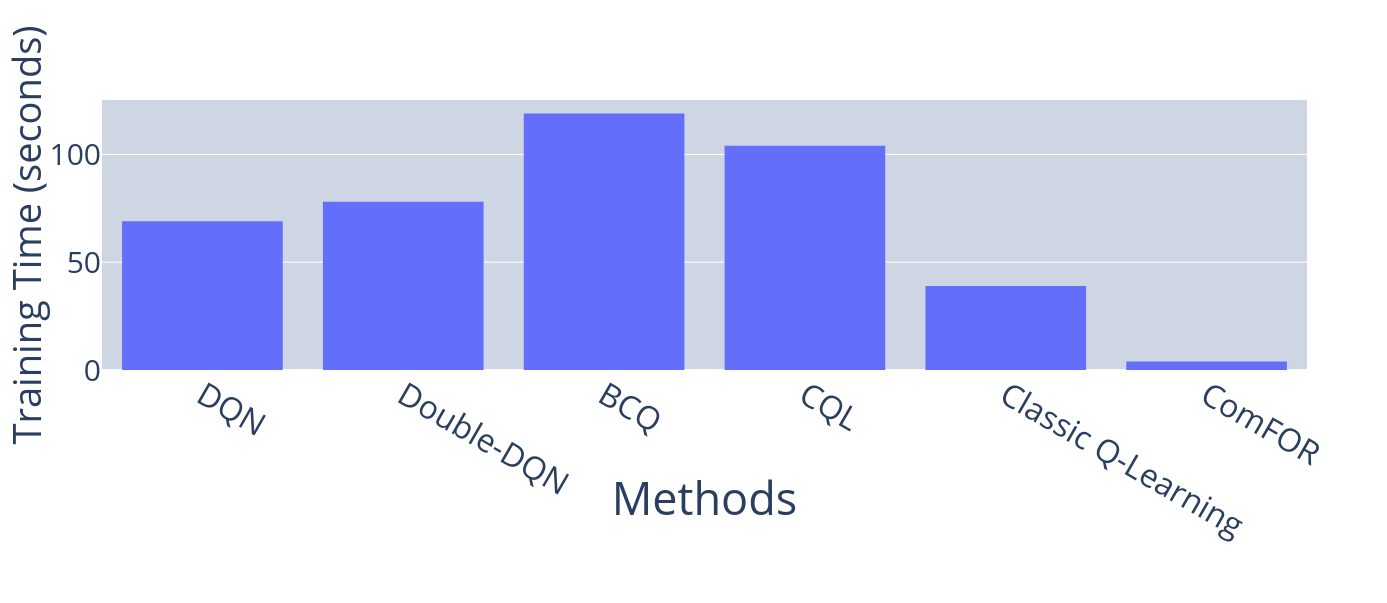}
		\caption{Training time on different methods.}
		\label{fig:Results_Performance}
	\end{figure}

	Finally, to compare performance and scalability, we present the training time for each method in Figure \ref{fig:Results_Performance}. All methods are executed on an AMD Ryzen Threadripper with 32 threads.
	
	Results from Figure \ref{fig:Results_Methods} show that our method ComFor outperforms other methods in both our proposed metric and the organizers' metric. Specifically, our method achieved a score that is 16\% higher than that of the state-of-the-art BCQ and 110\% higher than that of the baseline method provided by the organizers. Figure \ref{fig:Results_Performance} shows that ComFOR can train significantly faster than other methods, roughly 10 times faster than the classic Q-Learning method and 30 times faster than the state-of-the-art BCQ.
	
	\section{Conclusion and Future Work} 
	\label{sec:conclusion}
	
	In this study, we tackle the challenge of building a recommender system for a game that can be viewed as a multi-step decision-making problem. This problem also presents a difficulty for building recommender systems in the real world, which is the unavailability of a live training environment. The majority of today's recommender systems in e-commerce and retail businesses are based on supervised learning are built based on supervised learning models and collaborative filtering. In recent years, more researchers have proposed that deep reinforcement learning should be utilized for building recommender systems because they are well suited for addressing multi-item interactions, maximizing long-term rewards, and solving multi-step decision-making challenges. However, they suffer from the aforementioned problem of training on an offline data set. To tackle these challenges, we propose a novel reinforcement learning pipeline based on compressive features and clustering for assisting parallel Q-learning models. The results from our experiments show that our proposed method can train significantly faster than the other deep reinforcement learning methods and produce action policies that generate higher rewards.
	
	%% try targeted version
	%%%%%%%%%%%%%%%%%%%%%%%%%%%%%%%%%%%%%%%%%%%%%%%%%

	%------------------------------------------------------------------------------
	\bibliographystyle{IEEEtran}
	\bibliography{refs.bib}

\begin{thebibliography}{10}
\providecommand{\url}[1]{#1}
\csname url@samestyle\endcsname
\providecommand{\newblock}{\relax}
\providecommand{\bibinfo}[2]{#2}
\providecommand{\BIBentrySTDinterwordspacing}{\spaceskip=0pt\relax}
\providecommand{\BIBentryALTinterwordstretchfactor}{4}
\providecommand{\BIBentryALTinterwordspacing}{\spaceskip=\fontdimen2\font plus
\BIBentryALTinterwordstretchfactor\fontdimen3\font minus
  \fontdimen4\font\relax}
\providecommand{\BIBforeignlanguage}[2]{{%
\expandafter\ifx\csname l@#1\endcsname\relax
\typeout{** WARNING: IEEEtran.bst: No hyphenation pattern has been}%
\typeout{** loaded for the language `#1'. Using the pattern for}%
\typeout{** the default language instead.}%
\else
\language=\csname l@#1\endcsname
\fi
#2}}
\providecommand{\BIBdecl}{\relax}
\BIBdecl

\bibitem{ricci2015recommender}
F.~Ricci, L.~Rokach, and B.~Shapira, ``Recommender systems: introduction and
  challenges,'' in \emph{Recommender systems handbook}.\hskip 1em plus 0.5em
  minus 0.4em\relax Springer, 2015, pp. 1--34.

\bibitem{zhang2007agent}
T.~Zhang and D.~Zhang, ``Agent-based simulation of consumer purchase
  decision-making and the decoy effect,'' \emph{Journal of business research},
  vol.~60, no.~8, pp. 912--922, 2007.

\bibitem{zhu2014bundle}
T.~Zhu, P.~Harrington, J.~Li, and L.~Tang, ``Bundle recommendation in
  ecommerce,'' in \emph{Proceedings of the 37th international ACM SIGIR
  conference on Research \& development in information retrieval}, 2014, pp.
  657--666.

\bibitem{2021RL4RS}
K.~Wang, Z.~Zou, Q.~Deng, Y.~Shang, M.~Zhao, R.~Wu, X.~Shen, T.~Lyu, and
  C.~Fan, ``Rl4rs: A real-world benchmark for reinforcement learning based
  recommender system,'' \emph{ArXiv}, vol. abs/2110.11073, 2021.

\bibitem{richard1954theory}
B.~Richard, ``The theory of dynamic programming,'' \emph{Bulletin of the
  American Mathematical Society}, vol.~60, no.~6, pp. 503--516, 1954.

\bibitem{levine2020offline}
S.~Levine, A.~Kumar, G.~Tucker, and J.~Fu, ``Offline reinforcement learning:
  Tutorial, review, and perspectives on open problems,'' \emph{arXiv preprint
  arXiv:2005.01643}, 2020.

\bibitem{ma2021conservative}
Y.~J. Ma, D.~Jayaraman, and O.~Bastani, ``Conservative offline distributional
  reinforcement learning,'' \emph{arXiv preprint arXiv:2107.06106}, 2021.

\bibitem{van2016deep}
H.~Van~Hasselt, A.~Guez, and D.~Silver, ``Deep reinforcement learning with
  double q-learning,'' in \emph{Proceedings of the AAAI conference on
  artificial intelligence}, vol.~30, no.~1, 2016.

\bibitem{fujimoto2019off}
S.~Fujimoto, D.~Meger, and D.~Precup, ``Off-policy deep reinforcement learning
  without exploration,'' in \emph{International Conference on Machine
  Learning}.\hskip 1em plus 0.5em minus 0.4em\relax PMLR, 2019, pp. 2052--2062.

\bibitem{watkins1989learning}
C.~J. C.~H. Watkins, ``Learning from delayed rewards,'' 1989.

\bibitem{mnih2015human}
V.~Mnih, K.~Kavukcuoglu, D.~Silver, A.~A. Rusu, J.~Veness, M.~G. Bellemare,
  A.~Graves, M.~Riedmiller, A.~K. Fidjeland, G.~Ostrovski \emph{et~al.},
  ``Human-level control through deep reinforcement learning,'' \emph{nature},
  vol. 518, no. 7540, pp. 529--533, 2015.

\bibitem{lin1992self}
L.-J. Lin, ``Self-improving reactive agents based on reinforcement learning,
  planning and teaching,'' \emph{Machine learning}, vol.~8, no. 3-4, pp.
  293--321, 1992.

\bibitem{wei2021boosting}
H.~Wei, D.~Ye, Z.~Liu, H.~Wu, B.~Yuan, Q.~Fu, W.~Yang \emph{et~al.}, ``Boosting
  offline reinforcement learning with residual generative modeling,''
  \emph{arXiv preprint arXiv:2106.10411}, 2021.

\bibitem{fujimoto2019benchmarking}
S.~Fujimoto, E.~Conti, M.~Ghavamzadeh, and J.~Pineau, ``Benchmarking batch deep
  reinforcement learning algorithms,'' \emph{arXiv preprint arXiv:1910.01708},
  2019.

\bibitem{silver2014deterministic}
D.~Silver, G.~Lever, N.~Heess, T.~Degris, D.~Wierstra, and M.~Riedmiller,
  ``Deterministic policy gradient algorithms,'' in \emph{International
  conference on machine learning}.\hskip 1em plus 0.5em minus 0.4em\relax PMLR,
  2014, pp. 387--395.

\bibitem{kumar2020conservative}
A.~Kumar, A.~Zhou, G.~Tucker, and S.~Levine, ``Conservative q-learning for
  offline reinforcement learning,'' \emph{arXiv preprint arXiv:2006.04779},
  2020.

\bibitem{daylearning}
O.~Day, ``Learning autonomous robot grasping.''

\bibitem{sutton2018reinforcement}
R.~S. Sutton and A.~G. Barto, \emph{Reinforcement learning: An
  introduction}.\hskip 1em plus 0.5em minus 0.4em\relax MIT press, 2018.

\bibitem{krishnan2018challenges}
R.~Krishnan, D.~Liang, and M.~Hoffman, ``On the challenges of learning with
  inference networks on sparse, high-dimensional data,'' in \emph{International
  Conference on Artificial Intelligence and Statistics}.\hskip 1em plus 0.5em
  minus 0.4em\relax PMLR, 2018, pp. 143--151.

\bibitem{wold1987principal}
S.~Wold, K.~Esbensen, and P.~Geladi, ``Principal component analysis,''
  \emph{Chemometrics and intelligent laboratory systems}, vol.~2, no. 1-3, pp.
  37--52, 1987.

\bibitem{kramer1991nonlinear}
M.~A. Kramer, ``Nonlinear principal component analysis using autoassociative
  neural networks,'' \emph{AIChE journal}, vol.~37, no.~2, pp. 233--243, 1991.

\bibitem{johnstone2009consistency}
I.~M. Johnstone and A.~Y. Lu, ``On consistency and sparsity for principal
  components analysis in high dimensions,'' \emph{Journal of the American
  Statistical Association}, vol. 104, no. 486, pp. 682--693, 2009.

\bibitem{zou2006sparse}
H.~Zou, T.~Hastie, and R.~Tibshirani, ``Sparse principal component analysis,''
  \emph{Journal of computational and graphical statistics}, vol.~15, no.~2, pp.
  265--286, 2006.

\bibitem{makhzani2013k}
A.~Makhzani and B.~Frey, ``K-sparse autoencoders,'' \emph{arXiv preprint
  arXiv:1312.5663}, 2013.

\bibitem{bauckhage2014clustering}
C.~Bauckhage, A.~Drachen, and R.~Sifa, ``Clustering game behavior data,''
  \emph{IEEE Transactions on Computational Intelligence and AI in Games},
  vol.~7, no.~3, pp. 266--278, 2014.

\bibitem{hui2015spatial}
E.~C. Hui and C.~Liang, ``The spatial clustering investment behavior in housing
  markets,'' \emph{Land Use Policy}, vol.~42, pp. 7--16, 2015.

\bibitem{jin2021pessimism}
Y.~Jin, Z.~Yang, and Z.~Wang, ``Is pessimism provably efficient for offline
  rl?'' in \emph{International Conference on Machine Learning}.\hskip 1em plus
  0.5em minus 0.4em\relax PMLR, 2021, pp. 5084--5096.

\end{thebibliography}
	%\bibliography{Shared_Mendeley.bib}
	%------------------------------------------------------------------------------
	
	%------------------------------------------------------------------------------
	
	%------------------------------------------------------------------------------
\end{document}